\documentclass[runningheads]{llncs}
\usepackage[T1]{fontenc}
\usepackage{graphicx,verbatim}
\usepackage{algorithm}
\usepackage{algorithmic}
\usepackage{amsmath}

\usepackage{multirow}
\usepackage[table,xcdraw]{xcolor}
\usepackage{arydshln}
\usepackage{threeparttable}
\usepackage{colortbl}
\usepackage{makecell}

\usepackage{subcaption}

\begin{document}

\title{Can Diffusion Models Bridge the Domain Gap in Cardiac MR Imaging? }

\author{Xin Ci Wong \inst{1,2} \and Duygu Sarikaya \inst{2} \and Kieran Zucker \inst{3} \and Marc De Kamps \inst{2} \and Nishant Ravikumar \inst{2}}

\authorrunning{XC Wong et al.}

\institute{Centre for Doctoral Training in AI for Medical Diagnosis and Care, School of Computing, University of Leeds \email{scxcw@leeds.ac.uk} \and School of Computing, University of Leeds \and Leeds Cancer Centre, St James’s University Hospital, Leeds, UK}

\maketitle              
\begin{abstract}

    Magnetic resonance (MR) imaging, including cardiac MR, is prone to domain shift due to variations in imaging devices and acquisition protocols. This challenge limits the deployment of trained AI models in real-world scenarios, where performance degrades on unseen domains. Traditional solutions involve increasing the size of the dataset through ad-hoc image augmentation or additional online training/transfer learning, which have several limitations. Synthetic data offers a promising alternative, but anatomical/structural consistency constraints limit the effectiveness of generative models in creating image-label pairs. To address this, we propose a diffusion model (DM) trained on a source domain that generates synthetic cardiac MR images that resemble a given reference. The synthetic data maintains spatial and structural fidelity, ensuring similarity to the source domain and compatibility with the segmentation mask. We assess the utility of our generative approach in multi-centre cardiac MR segmentation, using the 2D nnU-Net, 3D nnU-Net and vanilla U-Net segmentation networks. We explore domain generalisation, where, domain-invariant segmentation models are trained on synthetic source domain data, and domain adaptation, where, we shift target domain data towards the source domain using the DM. Both strategies significantly improved segmentation performance on data from an unseen target domain, in terms of surface-based metrics (Welch’s t-test, $p < 0.01$), compared to training segmentation models on real data alone. The proposed method ameliorates the need for transfer learning or online training to address domain shift challenges in cardiac MR image analysis, especially useful in data-scarce settings.


\end{abstract}

\section{INTRODUCTION}

Cardiac magnetic resonance (MR) imaging has been extensively studied for automated segmentation \cite{Ahmad2023-gh,Ammar2021-xg}, but most models are trained and evaluated on the same dataset. In clinical practice, variations in imaging devices and acquisition protocols lead to domain shift, where differences in training and target data distribution degrade model performance on unseen data \cite{full2020studyingrobustnesssemanticsegmentation,Habijan2020-pa,Devran2022DomainShiftLVandRV}. To address domain shift, researchers have explored generative models such as GANs \cite{ALKHALIL2023106973,thermos2021controllablecardiacsynthesisdisentangled} to augment the data and Variational Autoencoders (VAE) to align feature distributions \cite{Cui2024VAE}. Domain-shift invariant CNN frameworks \cite{Patil2023-st} and soft-labeled contrastive learning \cite{10.1007/978-3-031-72114-4_65} have also been investigated for cardiac segmentation, along with variational approximation techniques \cite{Wu2021UDA}. However, the use of synthetic data as a direct substitute for real training or test data for downstream tasks remains underexplored, which is the focus of our work.

Diffusion models (DMs) \cite{ho2020denoising} known for generating high-quality images have recently been used for cardiac MR augmentation \cite{skorupko2024debiasingcardiacimagingcontrolled,Urcia2024DMCardiac}, but their potential to mitigate domain shift has not been thoroughly investigated. DMs have been observed to memorise and replicate training samples \cite{akbar2024bewarediffusionmodelssynthesizing}, raising questions about their effectiveness in generating useful synthetic medical data. Our approach is not limited by this because we integrate a reference image as guidance during the DM sampling process. Instead of generating images purely from noise, we iteratively refine the sampling process to ensure that the synthetic data remains structurally consistent with anatomical features while aligning with the source domain distribution. This distinguishes our approach from traditional DM-based augmentation, where uncontrolled synthesis may introduce undesired variations.

Recent work on Diffusion-Driven Adaptation (DDA) \cite{gao2023source} adapts target domain inputs back to the source domain using a DM trained on source data to mitigate input corruption, i.e. undesirable alterations in image appearance due to noise or artifacts. We extend this DDA by applying it to segmentation instead of classification, demonstrating its effectiveness in downstream pixel-wise tasks.

Our work focuses on using source domain-trained diffusion models (SD-DM) to address domain shift in medical imaging, particularly in data-scarce settings where collecting training data from multiple centres or vendors is infeasible. Our key contributions are: \textbf{(1) Diffusion-Driven Adaptation (DDA) for test-time input domain alignment:} we propose a simple yet effective strategy that leverages source domain-trained DM to adapt test-time inputs, ensuring that synthetic test images closely resemble real images while preserving anatomical structures. Unlike prior use of DDA, which primarily focus on data augmentation, our approach enables test-time domain alignment directly on the reference image for downstream medical image segmentation. \textbf{(2) Mixed adaptation for generalisable segmentation models:} we introduce a dual adaptation strategy, where both training and test data are adapted, enhancing segmentation robustness across domains without requiring explicit data augmentation.

\section{RELATED WORKS}

\textbf{Domain Adaptation (DA)} aims to align the source and target domains to reduce performance degradation. Most DA methods for medical image analysis fall into the single-source DA category and struggle with pixel-wise classification tasks \cite{Guan2022}. Our approach preserves pixel-level and structural similarities during adaptation. \textbf{Domain generalisation (DG)} enables models to generalise to unseen domains without knowledge of the target domain during training \cite{Yoon_2024}. Common DG strategies involve feature constraints and data augmentation \cite{huang2021fsdrfrequencyspacedomain}. However, GAN-based and adversarial approaches risk modifying domain-invariant features \cite{niemeijer2023generalizationadaptationdiffusionbaseddomain}, which can negatively impact downstream tasks. 
Furthermore, many DG methods also rely on auxiliary real data collected from different sources other than the target domain, whereas our method requires only a single training phase using synthetic data generated from the source domain.

\section{METHODS}

\begin{figure}
    \centering
    \includegraphics[width = 0.8\linewidth]{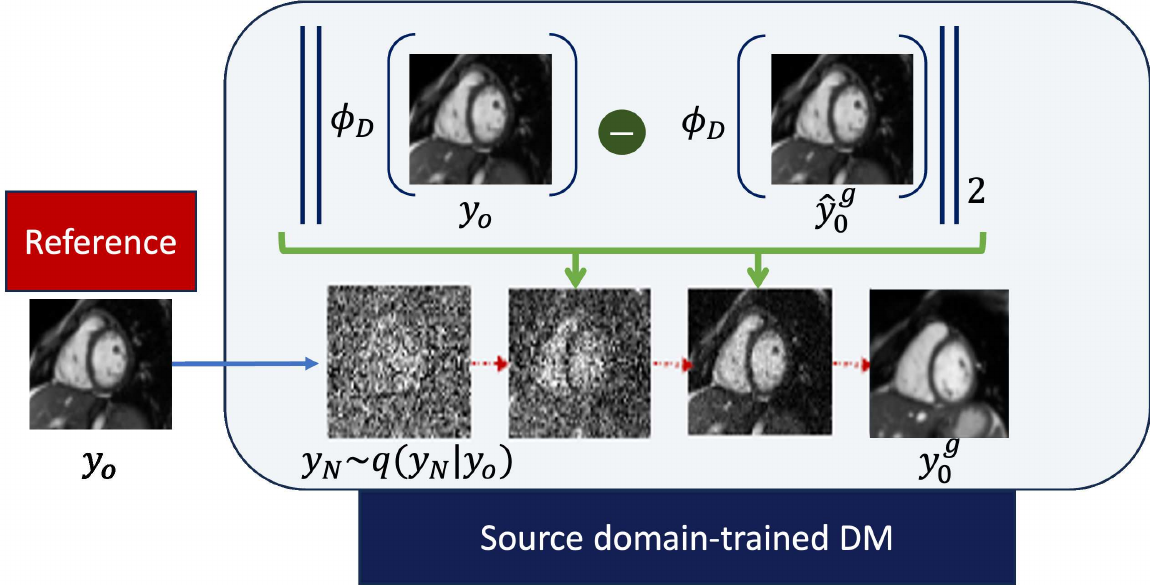}
   \caption{\textbf{The proposed reference-based cardiac MR synthesis using source domain-trained DM (SD-DM) can project target cardiac MR images to the source domain.} The process involves a forward diffusion (blue arrow) process that transforms the reference image through noise injection, followed by iterative denoising of the transformed input via a reverse diffusion (red arrow) process, while conditioning on the reference with a degree of scaling (denoted by the green arrows). The output $y_0^g$ is the adapted image which resembles the reference, $y_0$. The reference image can be sourced from any target domain.}
    \label{fig:reference-basedDM}
\end{figure}

In this section, we provide an overview of our approach, which leverages a source domain-trained diffusion model (SD-DM) to generate synthetic cardiac MR images aligned with the source domain (Fig. \ref{fig:reference-basedDM}). 

DM consists of a two-step process: a forward pass that adds noise to an image and a reverse pass that denoises it. During training, a neural network (U-Net in our approach) learns to predict and remove noise at different time steps, enabling image generation from pure noise. Our training and unconditional proposal are the same as Nichol \textit{et al.}\cite{nichol2021improveddenoisingdiffusionprobabilistic}. The unconditional proposal means the model is free to create from the learned distribution without specific guidance. For image synthesis, we use an iteratively refinement process \cite{gao2023source}, as detailed in Algorithm \ref{alg:diffusion_adaptation}. The process begins by corrupting the reference image $\mathbf{y}_0$ to a noisy version $\mathbf{y}_N$ through a forward diffusion process. Starting from this noisy initialisation $\mathbf{y}_N^g$, the algorithm performs a reverse denoising process across $N$ steps, progressively generating samples closer to the source domain distribution. At each timestep $t$, an intermediate denoised sample $\hat{\mathbf{y}}_{t-1}^g$ is obtained through the learned reverse process. A predicted clean image $\hat{\mathbf{y}}_0^g$ is then computed using the standard DDPM reparameterisation formula. We incorporate a linear low-pass filter $\phi_D(\cdot)$ that provides a sequence of downsampling and upsampling operation with scale factor of $D$ which determines the image-level structure refinement \cite{choi2021ilvrconditioningmethoddenoising,gao2023source}. The gradient $\nabla{\mathbf{y}_t}$ minimises the L$2$ distance between the filtered features of the reference and generated images. The guidance strength $\tau$ controls the trade-off between maintaining fidelity to the original reference and introducing randomness. The update of $\mathbf{y}_t^g$ now has structural similarity to the reference image in scale $D$. After $N$ steps, the final adapted image $\mathbf{y}_0^g$ preserves structural content from the original input while reflecting the visual characteristics of the source domain. 

Both source and target domain are considered semantically homogeneous for cardiac MR datasets, i.e. the anatomical structures are consistent across both the source ($x_s$) and target ($x_t$) domains. Due to the variation caused by differences in vendors and imaging protocols, the intensity distributions differ, ($p(x_t) \neq p(x_s)$). For DG, we enforce $p(f(x_s)) = p(x_s)$, where $f(\cdot)$ represents the diffusion model. Webber \textit{et al.} \cite{Webber2024-in} highlighted the advantage of using DM for medical image reconstruction for its ability to decouple the image prior, i.e. a learned representation of the expected characteristics of a typical MR image, from the scanner parameters. This assumes that the DM suppresses domain-specific artifacts, such as scanner-induced intensity variations, while generating images that resemble themselves. This approach aligns with domain-invariant representation learning, a sub-category of DG, where the DM removes spurious domain-related variations, and this synthetic data is then used for downstream segmentation training, improving generalisation across unseen domains \cite{wang2022generalizingunseendomainssurvey}.

Conversely, for DA, we replace the reference for SD-DM with target domain image to generate its synthetic counterpart. This test-time adaptation (TTA) ensures that the transformed target image follow the source domain distribution, achieving $p(y|f(x_t)) = p(y|x_s)$ \cite{Liang_2024}. Since the same annotation protocol was used for both source and test datasets, we assume no concept shift ($p(y | f(x_s)) = p(y | x_t)$) occurs, that is, the probability of a given pixel belonging to a certain anatomical class is still the same. However, label shift ($p(y_s) \neq p(y_t)$) due to class imbalance is expected to persist.

We assess our synthetic data effectiveness for domain shift with downstream segmentation tasks using 2D nnU-Net, 3D nnU-Net and vanilla U-Net \cite{isensee2024nnunetrevisitedrigorousvalidation,ronneberger2015unet}. 

\begin{algorithm}[!htbp]
    \caption{Iterative Refinement for Sampling}
    \label{alg:diffusion_adaptation}
    \begin{algorithmic}[1]
        \STATE \textbf{Input:} Source domain-trained diffusion model  $g(\mathbf{y},w,t)$, Reference image $\mathbf{y}_0$
        \STATE \textbf{Hyperparameter:} N: Diffusion range, $\phi_D(\cdot)$: Low pass filter of $D$, $\tau$: Controlling guidance strength, $\beta_t$: Noise schedule
        \STATE \textbf{Output:} Domain-adapted image $\mathbf{y}_0^g$ 
        \STATE Initialise diffusion model with parameter $w$
        \STATE Sample $\mathbf{y}_N \sim q(\mathbf{\mathbf{y}_N}|\mathbf{y}_0)$    // Corrupt reference image $\mathbf{y}_0$ until $\mathbf{y}_N$
        \STATE $\mathbf{y}_N^g \leftarrow \mathbf{y}_N$ 
        \FOR{$t \leftarrow N \ldots 1$} 
            \STATE $\hat{\mathbf{y}}_{t-1}^g \sim p_\theta(\mathbf{y}_{t-1}^g | \mathbf{y}_{t}^g)$ // Unconditional proposal
            \STATE $\hat{\mathbf{y}}_0^g \leftarrow \sqrt{\frac{1}{\bar\alpha_t}}\mathbf{y}_t^g - \sqrt{\frac{1}{\bar{\alpha}_t}-1} g(\mathbf{y}_t^g,w, t)$ // $\alpha_t := 1 - \beta_t, \quad \bar{\alpha}_t := \prod_{s=1}^{t} \alpha_s$
            \STATE $\mathbf{y}_{t-1}^g \leftarrow \hat{\mathbf{y}}_{t-1}^g - \tau\nabla_{\mathbf{y}_t}\parallel \phi_D(\mathbf{y}_0) - \phi_D(\hat{\mathbf{y}}_{0}^g) \parallel_2$
        \ENDFOR
        \RETURN $\mathbf{y}_0^g$
    
    \end{algorithmic}
\end{algorithm}

\section{EXPERIMENTS}

\begin{figure}[ht]
    \centering
    \includegraphics[width = 0.70\linewidth]{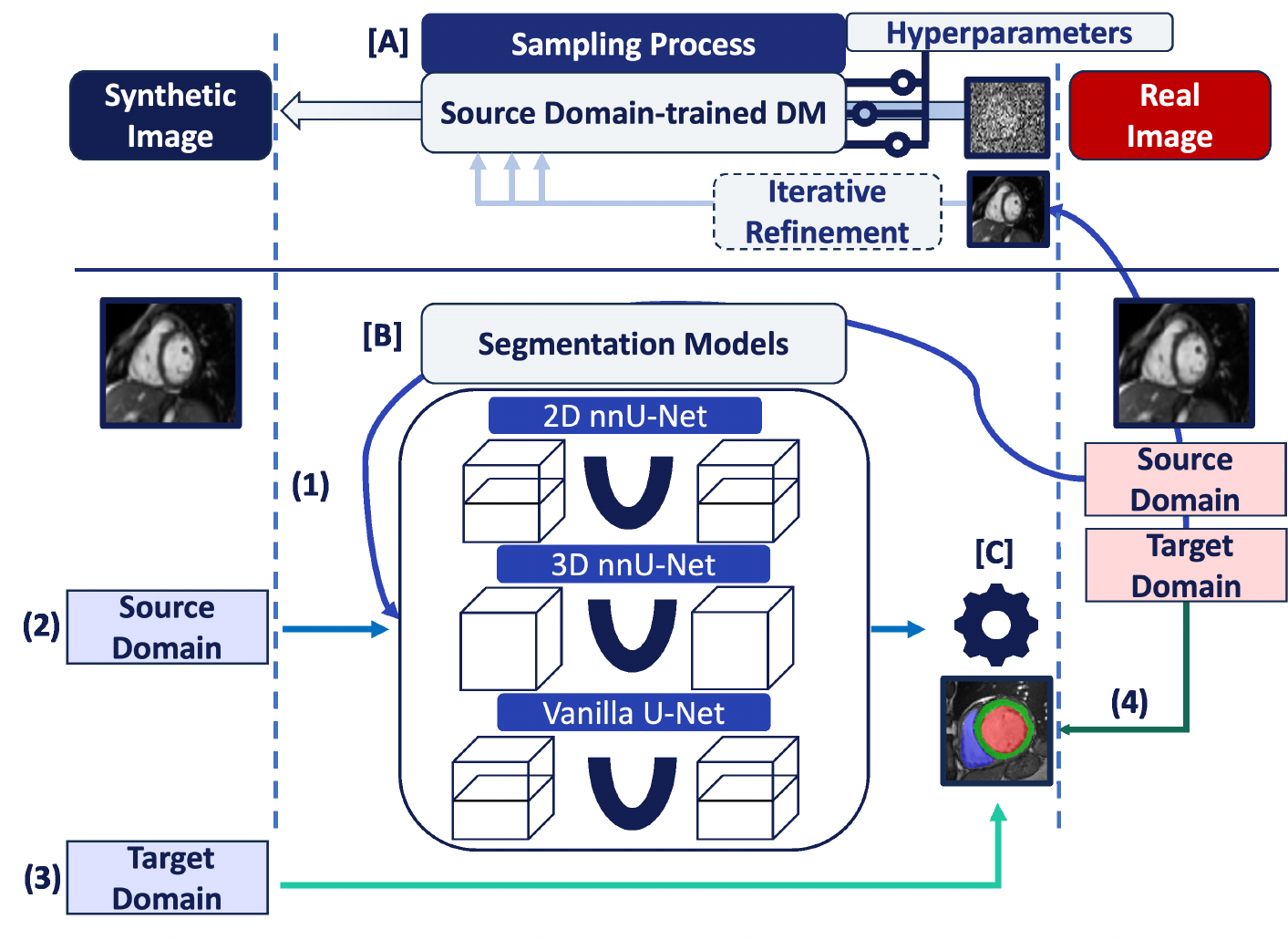}
   \caption{\textbf{Overview of the Experimental Design.} \textbf{[A]} In the sampling process, the real image from either the source domain (M\&Ms) or the target domain (ACDC) is selected as reference. The trained DM has hyperparameters: scale factor $D$, diffusion range $N$, and controlling
guidance strength $\tau$, to refine the generative process and enhance similarity to the reference image. \textbf{[B]} For segmentation, we evaluate three models: 2D nnU-Net, 3D nnU-Net, and Vanilla U-Net, each trained on either (1) real source domain data or (2) synthetic source domain data. \textbf{[C]} During prediction, models are tested separately on (3) the synthetic target domain dataset and (4) the original target domain dataset.}
    \label{fig:designflow}
\end{figure}

\subsubsection{Datasets} These experiments were done using Multi-Centre, Multi-Vendor \& Multi-Disease Cardiac Image Segmentation Challenge (M\&Ms) dataset \cite{Campello2021-ar} as the source domain and the Automatic Cardiac Diagnosis Challenge (ACDC) dataset \cite{Bernard2018-au} as the target domain. Both short-axis cardiac MR datasets share the same annotation protocol, labeling three main structural regions: left ventricle (LV), right ventricle (RV), and left ventricle myocardium (MYO). We focus on end-diastolic and end-systolic frames, following the original M\&Ms data split, using $3284$ slices ($300$ scans) to train both the DM and segmentation models.

\subsubsection{Segmentation} We trained the 2D nnU-Net, 3D nnU-Net and vanilla U-Net (Fig. \ref{fig:designflow}, Detailed U-Net at Appendix \ref{app:unet}) for 250 epochs each without data augmentation and without cross-validation to ensure fair evaluation. Ablation studies were conducted on real ACDC ($2978$ slices, $300$ scans), domain-adapted ACDC ($D=2$, $N=25$)\footnote{Synthetic data generated with scale factor $D$ and diffusion range $N$, with fixed $\tau = 6$}, and the reserved M\&Ms test set ($3236$ slices, $272$ scans). DG models were trained using synthetic M\&Ms ($D=2$, $N=25$), while mixed adaptation evaluated synthetic ACDC on synthetic M\&Ms-trained segmentation models.

\subsubsection{Implementation details} We clipped reference image intensity values to the $0.5$th to $99.5$th percentile before scaling to $[0,255]$ \cite{Ma2024-ca}. Our DM shared the sampling method and architecture of the improved DM \cite{nichol2021improveddenoisingdiffusionprobabilistic}, processing images at resolutions of 
$128 \times 128$ and $256 \times 256$ pixels. We experimented with diffusion steps of $1000$, $2000$ and $3000$, using either a linear or cosine noise scheduler. The model is configured to learn the noise scale parameter, $\sigma$ for fast sampling. The generated output is projected back into NIfTI format to maintain compatibility with the original dataset. We used NVIDIA's L$40$S GPU for all experiments. The code is available at: https://github.com/X-ksana/sddm

\section{RESULTS \& DISCUSSION}

We evaluated our approach using both qualitative and quantitative assessments. Qualitative analysis was based on visual inspection of synthetic images, while quantitative performance was measured using segmentation metrics \cite{müller2022guidelineevaluationmetricsmedical}, computed with MedPy\footnote{https://github.com/loli/medpy}. Additionally, we analysed the feature distribution using t-SNE visualisation to assess domain alignment. These evaluations helped determine whether our synthetic data can replicate the effects of real data.

\subsubsection{Structural Consistency} 

\begin{figure}[!htbp]
    \centering
    \begin{minipage}[b]{0.48\linewidth}
        \centering
        \includegraphics[width=\linewidth]{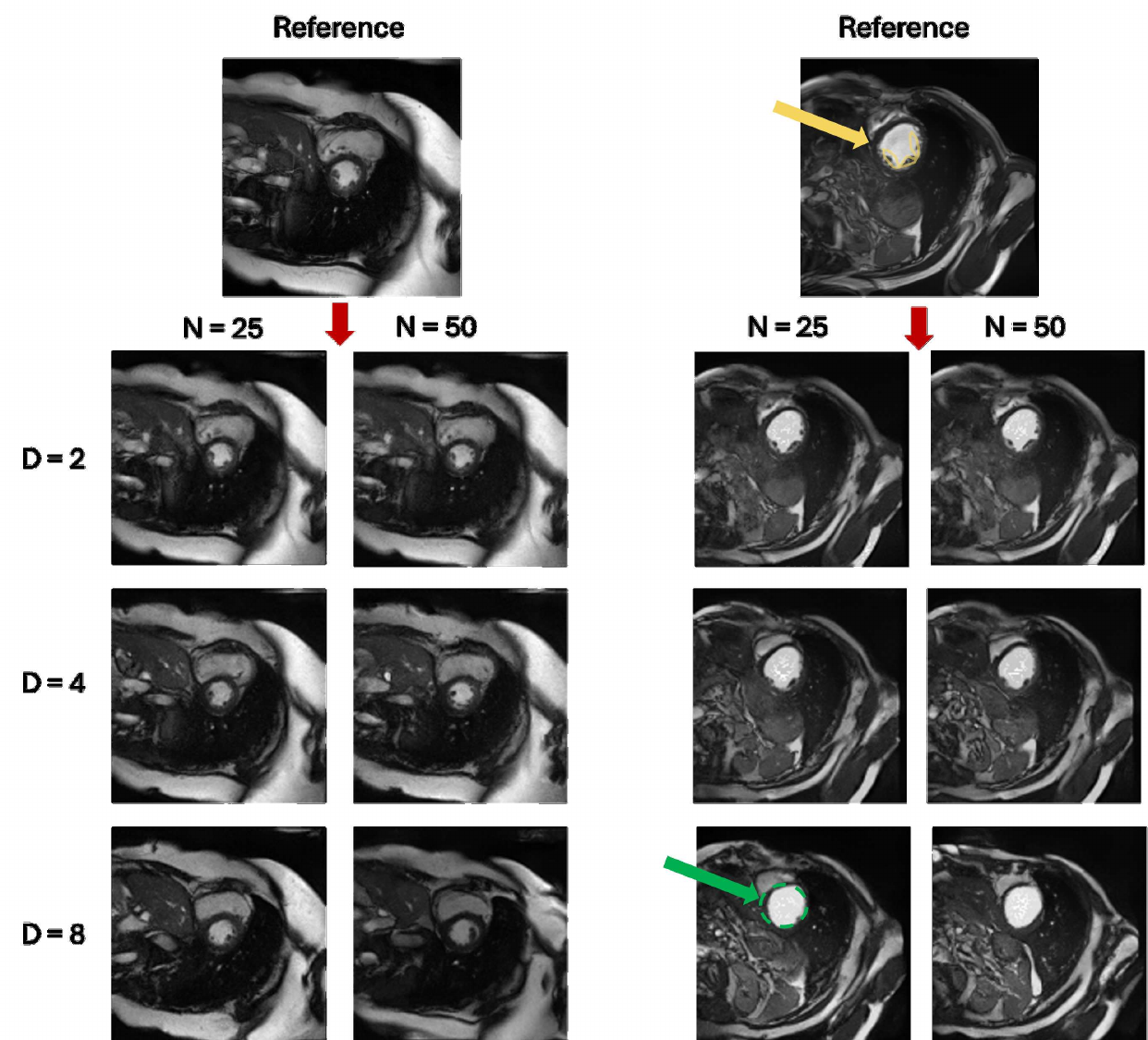}
        \caption{\textbf{Image synthesis using various scaling factor $D$ and diffusion range $N$ with fixed $\tau = 6$.} As $D$ increases, more semantic information lost (yellow $\rightarrow$ green region).}
        \label{fig:ablate_d}
    \end{minipage}
    \hfill
    \begin{minipage}[b]{0.48\linewidth}
        \centering
        \includegraphics[width=\linewidth]{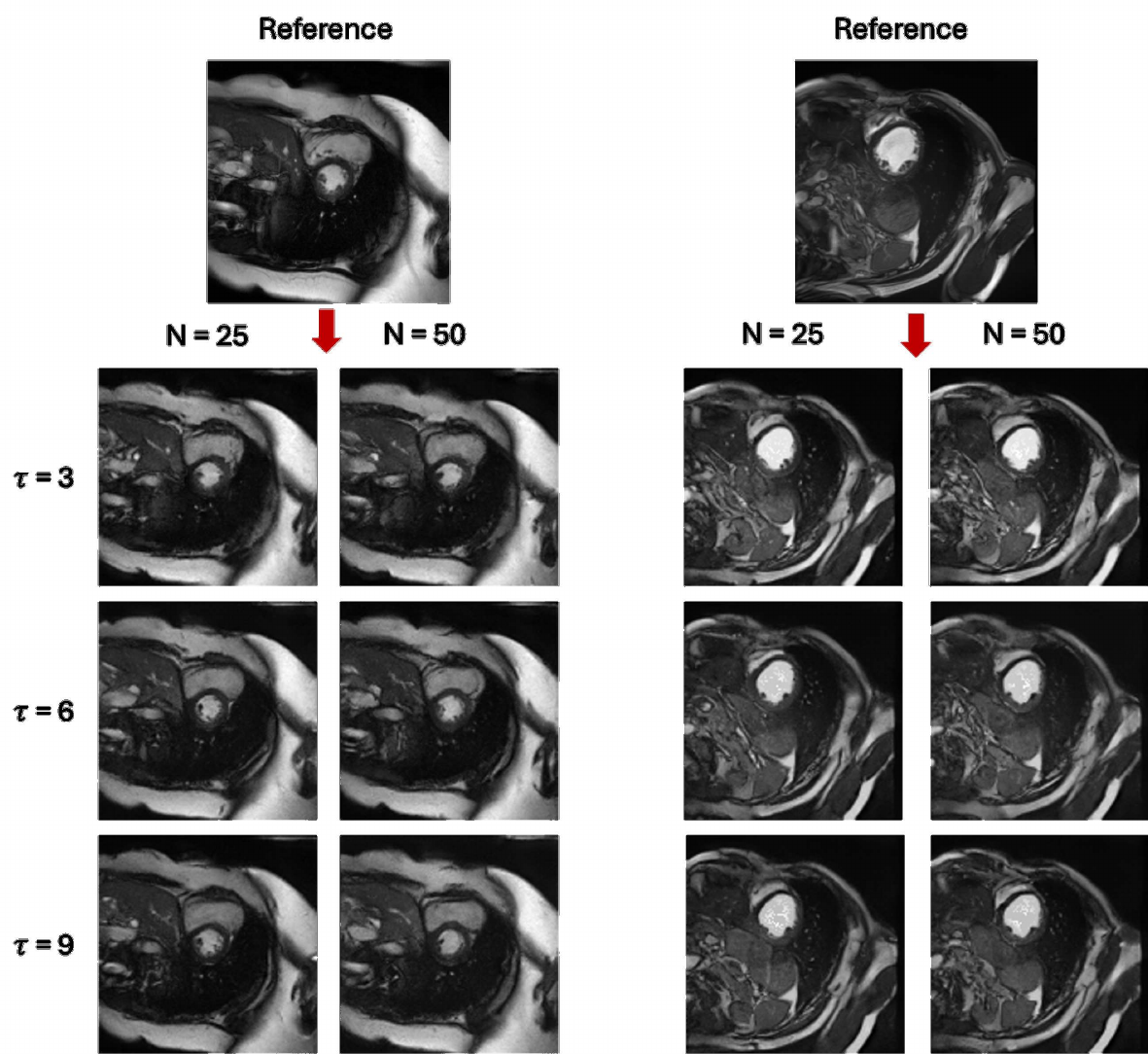}
        \caption{\textbf{Image synthesis using various $\tau$ and diffusion range $N$ with fixed $D = 4$.} Since $\tau$ controls the step size of guidance, a very small $\tau$ results in no guidance.}
        \label{fig:ablate_tau}
    \end{minipage}
\end{figure}

\begin{figure}[!htbp]
    \centering
    \begin{minipage}{0.2\textwidth}
        \centering
        \includegraphics[width=\linewidth]{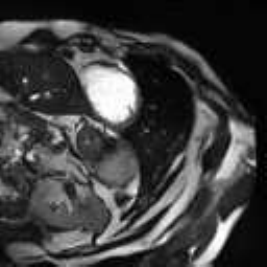} 
    \end{minipage}\hfill
    \begin{minipage}{0.2\textwidth}
        \centering
        \includegraphics[width=\linewidth]{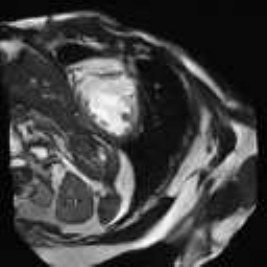} 
    \end{minipage}\hfill
    \begin{minipage}{0.2\textwidth}
        \centering
        \includegraphics[width=\linewidth]{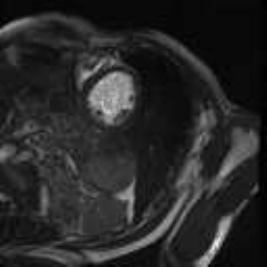} 
    \end{minipage}\hfill
    \begin{minipage}{0.2\textwidth}
        \centering
        \includegraphics[width=\linewidth]{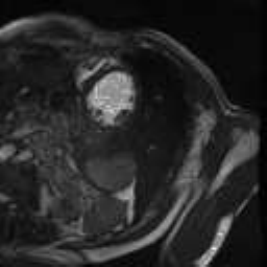} 
    \end{minipage}
    \caption{\textbf{Controlling structural consistency by extremal settings of $D$ and $\tau$.} All images are synthesised using the same reference (indicated with a yellow arrow in Fig. ~\ref{fig:ablate_d}). From left to right: ($D=16$, $\tau=1$), ($D=32$, $\tau=1$), ($D=1$, $\tau=18$), and ($D=1$, $\tau=24$). Large $D$ induces semantic drift; high $\tau$ leads to more preservation of the reference anatomy.}
    \label{fig:control}
\end{figure}

Synthetic cardiac MR images (Fig. \ref{fig:ablate_d} and Fig. \ref{fig:ablate_tau}) closely resembled their respective reference image with good reconstruction on the three important regions of interest (LV, RV, MYO). With different refinement scale for image adaptation, the cardiac MR images had different level of reconstruction, highlighting a controllable trade-off between introducing novel synthetic structures and preserving reference anatomy (Fig. \ref{fig:control}). Visual assessment revealed occasional white artifacts in the background of the generated images, likely caused by residual noise retained during the reverse process. Nevertheless, the overall anatomical structures remained well-preserved, even in the presence of a limited dataset. While diffusion models are traditionally considered data-hungry, our results suggest that structural fidelity can be maintained with constrained data availability.

\subsubsection{Domain Alignment} 
\begin{figure}
    \centering
    \includegraphics[width =\linewidth]{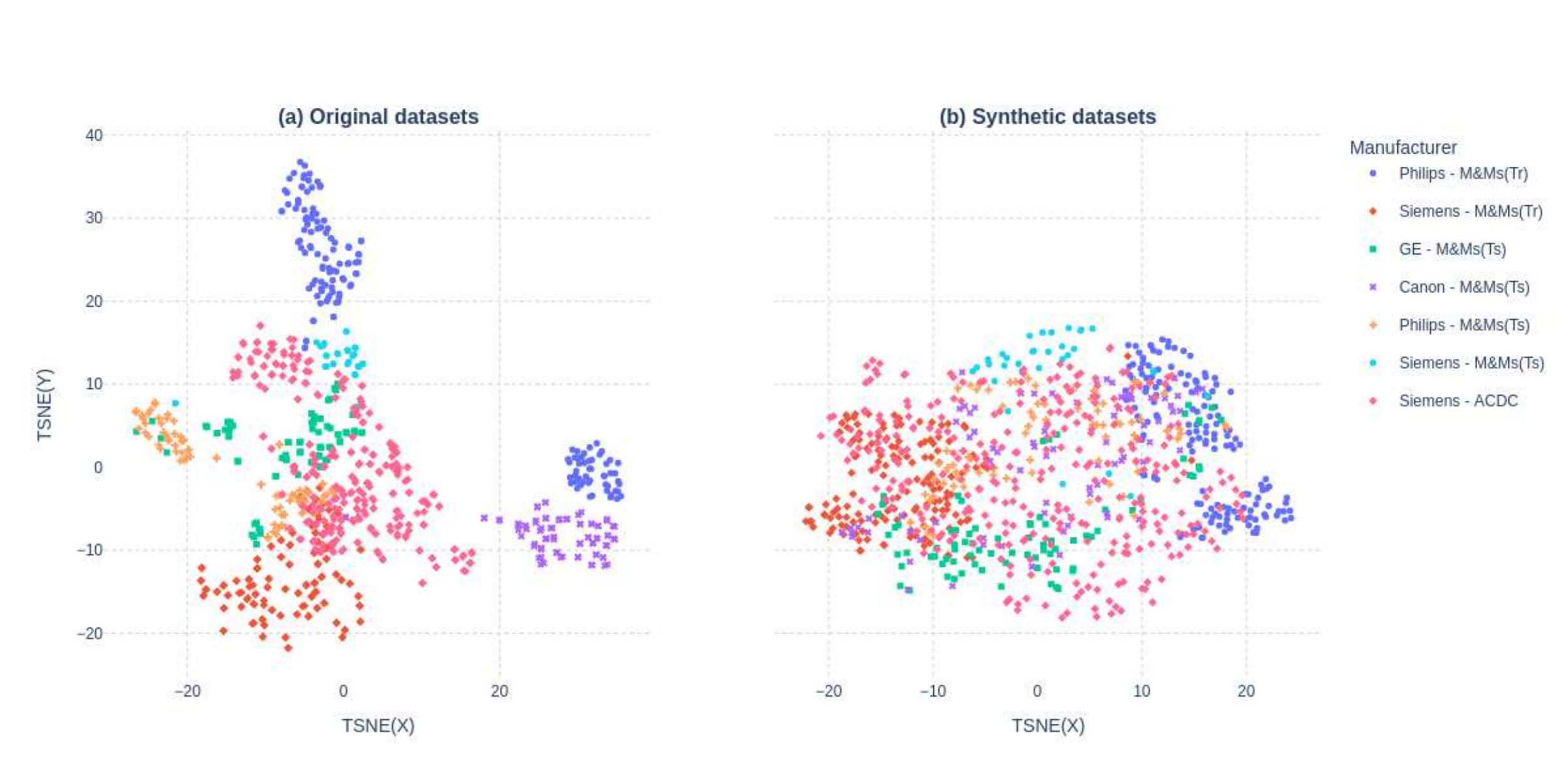}
   \caption{\textbf{t-SNE visualisation of (a) original datasets and (b) synthetic datasets.} Each point represents a cardiac MR scan. (a) Distinct clustering patterns are observed for different manufacturers, indicating domain shifts. (b) The synthetic data exhibits improved overlap, forming a merged cluster which suggests enhanced domain alignment and reduced domain-specific variation.}
    \label{fig:tsne_plot}
\end{figure}

t-SNE\footnote{t-distributed stochastic neighbor embedding,  https://github.com/rkushol/DSMRI} converted $22$ domain-related features in each cardiac MR into a two-dimensional space using default setting (n\_components$= 2$, perplexity$= 30$)\cite{kushol2023dsmri}. The t-SNE plots of the original and synthetic datasets (Fig. \ref{fig:tsne_plot}) reveal distinct shifts, highlighting the influence of acquisition settings and manufacturers on the data distribution. Our SD-DM can reduce the distribution gap between the datasets as the representations now became more indistinguishable. 

\begin{table}[!ht]
\centering
\begin{threeparttable}
\caption{\textbf{Results for Ablation Experiments.} Models of the same type are grouped using different superscript symbols. \textbf{Bolded} values indicate the best results in each group, while {\color[HTML]{00009B}\underline{blue underlined}} values are not significantly lower than the best (Welch’s t-test, p-value $> 0.01$). Surface-based metrics are reported in millimeters.}
\label{tab:performance}
\centering
\begin{tabular}{l|l|cc|cccc}
\hline
\multicolumn{1}{c|}{\multirow{2}{*}{\textbf{Approach}}} & \multicolumn{1}{c|}{\multirow{2}{*}{\textbf{Model}}} & \multicolumn{2}{c|}{\textbf{Overlap-based}} & \multicolumn{4}{c}{\textbf{Surface-based}} \\ 
\cline{3-8}
\multicolumn{1}{c|}{} & \multicolumn{1}{c|}{} & DSC ↑ & IoU ↑ & ASD ↓ & ASSD ↓ & HD ↓ & HD95 ↓ \\ 
\hline
\multirow{3}{*}{\makecell[l]{Baseline with \\ Real ACDC}} 
    & 2D nnU-Net\textsuperscript{†} & \textbf{0.839} & \textbf{0.736} & \color[HTML]{00009B}\underline{1.6} & 1.6 & \color[HTML]{00009B}\underline{24.0} & 6.7 \\
    & 3D nnU-Net\textsuperscript{‡} & \textbf{0.810} & \textbf{0.697} & 3.8 & 3.0 & 53.3 & 13.4 \\
    & 2D U-Net\textsuperscript{§} & \textbf{0.859} & \textbf{0.762} & 1.6 & 1.4 & 36.5 & 6.9 \\
\hline
\multirow{3}{*}{\makecell[l]{DA with \\ Synthetic ACDC}} 
    & 2D nnU-Net\textsuperscript{†} & 0.792 & 0.678 & \textbf{1.2} & \textbf{1.1} & \color[HTML]{00009B}\underline{23.8} & \color[HTML]{00009B}\underline{4.9} \\
    & 3D nnU-Net\textsuperscript{‡} & 0.757 & 0.638 & 4.8 & 3.2 & 52.9 & 13.0 \\
    & 2D U-Net\textsuperscript{§} & \color[HTML]{00009B}\underline{0.854} & \color[HTML]{00009B}\underline{0.754} & \textbf{0.7} & \textbf{0.6} & 33.8 & \textbf{2.4} \\
\hline
\multirow{3}{*}{\makecell[l]{DG with \\ Real ACDC}} 
    & 2D nnUnetSyn\textsuperscript{†} & 0.656 & 0.520 & 3.0 & 3.6 & 33.9 & 12.1 \\
    & 3D nnUnetSyn\textsuperscript{‡} & 0.688 & 0.552 & 5.2 & 4.4 & 62.7 & 17.2 \\
    & 2D UnetSyn\textsuperscript{§} & 0.705 & 0.566 & 2.2 & 2.9 & 34.7 & 10.0 \\
\hline
\multirow{3}{*}{\makecell[l]{Mixed Adaptation \\ with Synthetic ACDC}} 
    & 2D nnUnetSyn\textsuperscript{†} & 0.736 & 0.604 & \color[HTML]{00009B}\underline{1.4} & \color[HTML]{00009B}\underline{1.3} & \textbf{23.4} & \textbf{4.7} \\
    & 3D nnUnetSyn\textsuperscript{‡} & 0.735 & 0.601 & \textbf{2.0} & \textbf{1.6} & \textbf{36.2} & \textbf{6.1} \\
    & 2D UnetSyn\textsuperscript{§} & 0.750 & 0.618 & 0.9 & 0.9 & \textbf{26.1} & \color[HTML]{00009B}\underline{3.2} \\
\hline
\end{tabular}
\begin{tablenotes}
\small
\item [1] DSC: Dice Similarity Coefficient; IoU: Intersection-over-Union; ASD: Average Surface Distance; ASSD: Average Symmetric Surface Distance; HD: Hausdorff Distance; HD95: 95\% Hausdorff Distance. DA: Domain Adaptation (test-time input adaptation on target data); DG: Domain Generalisation (domain-invariant models trained on synthetic source data), \{model name\}Syn denotes models trained on synthetic M\&Ms data (e.g., 2D UnetSyn).
\end{tablenotes}
\end{threeparttable}
\end{table}

\subsubsection{Segmentation Performance} We evaluated our approach using 2D nnU-Net, 3D nnU-Net, and vanilla U-Net, comparing their performance on real and synthetic datasets (Table \ref{tab:performance}). Performance metrics with 95\% confidence intervals across anatomical structures are provided in Appendix \ref{app:graph}. Models trained on synthetic data demonstrated statistically significant improvements in surface-based metrics (Welch’s t-test, $P < 0.01$) compared to those trained on real datasets, suggesting smoother predicted masks with reduced variability. Traditionally, such performance requires large annotated datasets, advanced architectures, or extensive computational resources \cite{Chen2020-uz}. However, in our case, these gains were achieved with limited annotations and without data augmentation, indicating the potential of this method. However, overlap-based metrics showed modest declines or no significant differences from the baseline. This align with findings from prior work, where the usage of synthetic brain MRIs for segmentation led to minimal improvement or declines due to the strength of the baseline models, which are already highly optimised and difficult to surpass \cite{Usman_Akbar2024-cg}.
We also observed occasional instability in synthetic images, likely due to residual noise from the fast sampling process. This led to scattered segmentations and a decline in overlap-based metrics as well. It is possible that with augmentation the results will improve. However, it would confound our aim of demonstrating the standalone utility of the alignment feature of the diffusion models. Future work could adopt ensembling strategies to overcome instability, as explored by Gao \textit{et al.} \cite{gao2023source}, who combine synthetic and original images in classification output to compensate for synthetic image variability. Post-processing techniques that retain only the largest connected component in the predicted mask also could help in removing scattered segmentations. Additionally, we can use better sampling techniques such as FastDPM \cite{kong2021fastsamplingdiffusionprobabilistic} for improved fast sampling quality without retraining, or noise level correction \cite{abuduweili2025enhancingsamplegenerationdiffusion}, which refines noise level estimates during the denoising process. 

\section{CONCLUSION}

Our SD-DM approach enables mixed adaptation by combining test-time adapted target data with a synthetic source-trained segmentation model, achieving robust performance across sites and vendors without fine-tuning or data augmentation, making it ideal for data-scarce settings. t-SNE visualisation shows that SD-DM aligns feature distributions, promoting a shared latent space. By leveraging anatomically consistent synthetic data while mitigating domain shift, our method achieves segmentation performance comparable to real datasets, reducing reliance on costly data collection and manual annotation. However, a trade-off remains between fast sampling and image quality, where artifacts can affect fine structural details. Future work will explore ensemble-based fusion of real and synthetic data to enhance segmentation robustness.

\clearpage

\begin{credits}
\subsubsection{\ackname} We acknowledge the support of the Centre for Doctoral Training in AI for Medical Diagnosis and Care for funding this project. We extend our gratitude to the patients and doctors who contributed to datasets. The study was facilitated using the Aire High Performance Computing resources at the University of Leeds, UK. 

\subsubsection{\discintname}
All authors have no competing interests to declare that are relevant to the content of this article.
\end{credits}

%
%
%

\bibliographystyle{splncs04}
\bibliography{paper}

\appendix
\section{Appendix A: Vanilla U-Net Architecture}\label{app:unet}

We implemented a 2D vanilla U-Net model in PyTorch to serve as a baseline segmentation architecture. The network follows the original U-Net design by Ronneberger \textit{et al.}, with modifications including batch normalisation after each convolution layer to improve training stability \cite{ronneberger2015unet}.

The architecture consists of:

\begin{itemize}
    \item \textbf{Encoder:} Four downsampling blocks, each composed of two $3\times3$ convolutional layers, batch normalisation, and ReLU activation. Each block is followed by a max pooling with a stride of $2$.
    
    \item \textbf{Bottleneck:} A double convolution block with 16$\times$ the base number of filters, serving as the bridge between encoder and decoder.
    
    \item \textbf{Decoder:} Four upsampling blocks, each consisting of a transposed convolution layer (stride 2, kernel size 2) for upsampling, followed by concatenation with the corresponding encoder feature map, and a double convolution block.
    
    \item \textbf{Output:} A $1\times1$ convolution maps the decoder output to the desired number of segmentation classes.
\end{itemize}

The model was initialised with a base filter size of 64 and trained using 1-channel grayscale input images with 4 output classes (background, LV, MYO, RV).

The U-Net served as a lightweight reference for benchmark segmentation performance under limited data and without augmentation, complementing the more advanced nnU-Net configurations used in the main experiments.

\section{Appendix B: Quantitative Performance Metrics with Confidence Intervals}\label{app:graph}

\begin{figure}[htb]
    \centering
    \begin{subfigure}[b]{0.48\linewidth}
        \centering
        \includegraphics[width=\linewidth]{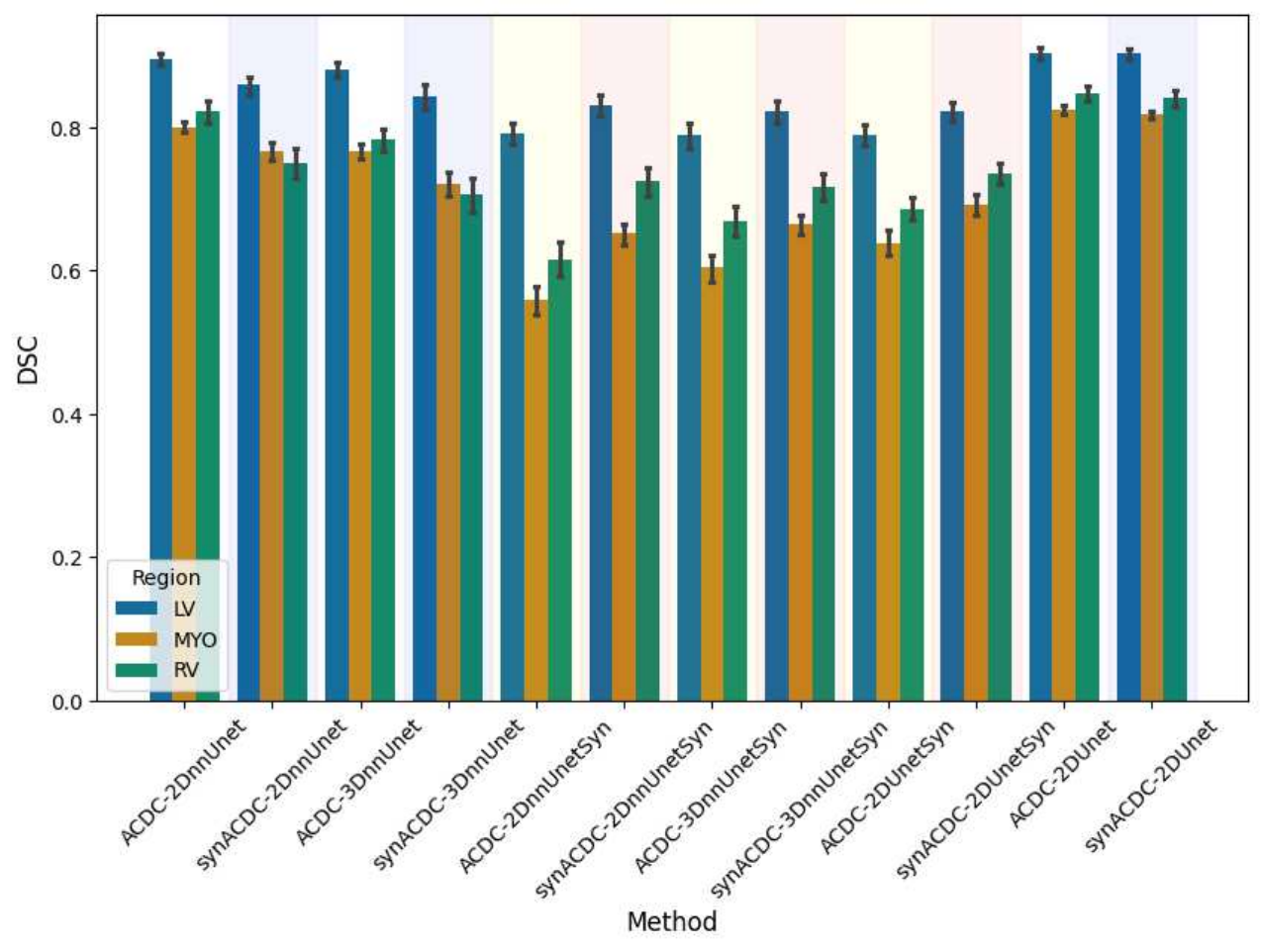}
        \caption{\textbf{Mean DSC}}
        \label{fig:dice_plot}
    \end{subfigure}
    \hfill
    \begin{subfigure}[b]{0.48\linewidth}
        \centering
        \includegraphics[width=\linewidth]{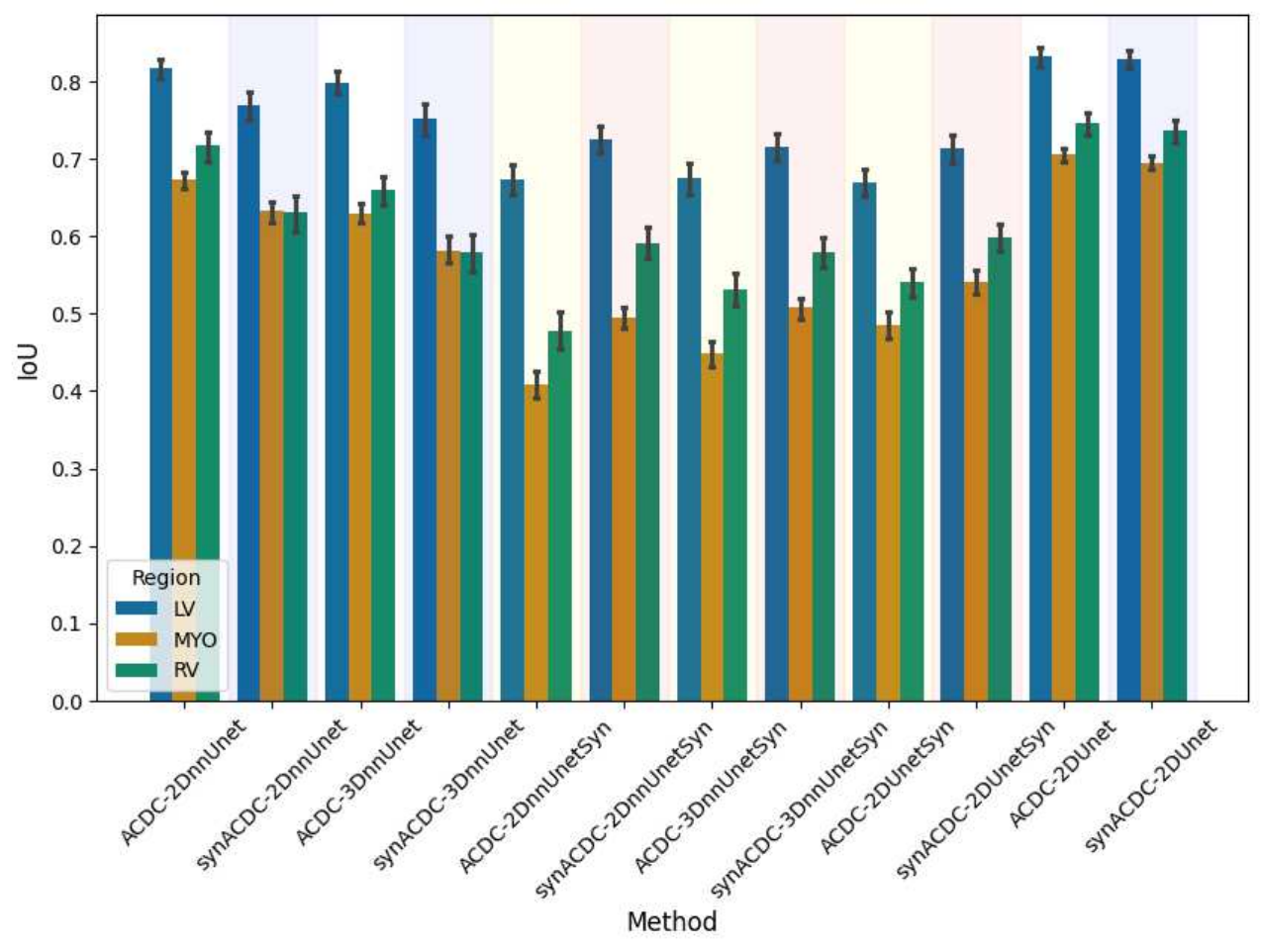}
        \caption{\textbf{Mean IoU}}
        \label{fig:iou_plot}
    \end{subfigure}
    
    \vspace{0.5em}
    
    \begin{subfigure}[b]{0.48\linewidth}
        \centering
        \includegraphics[width=\linewidth]{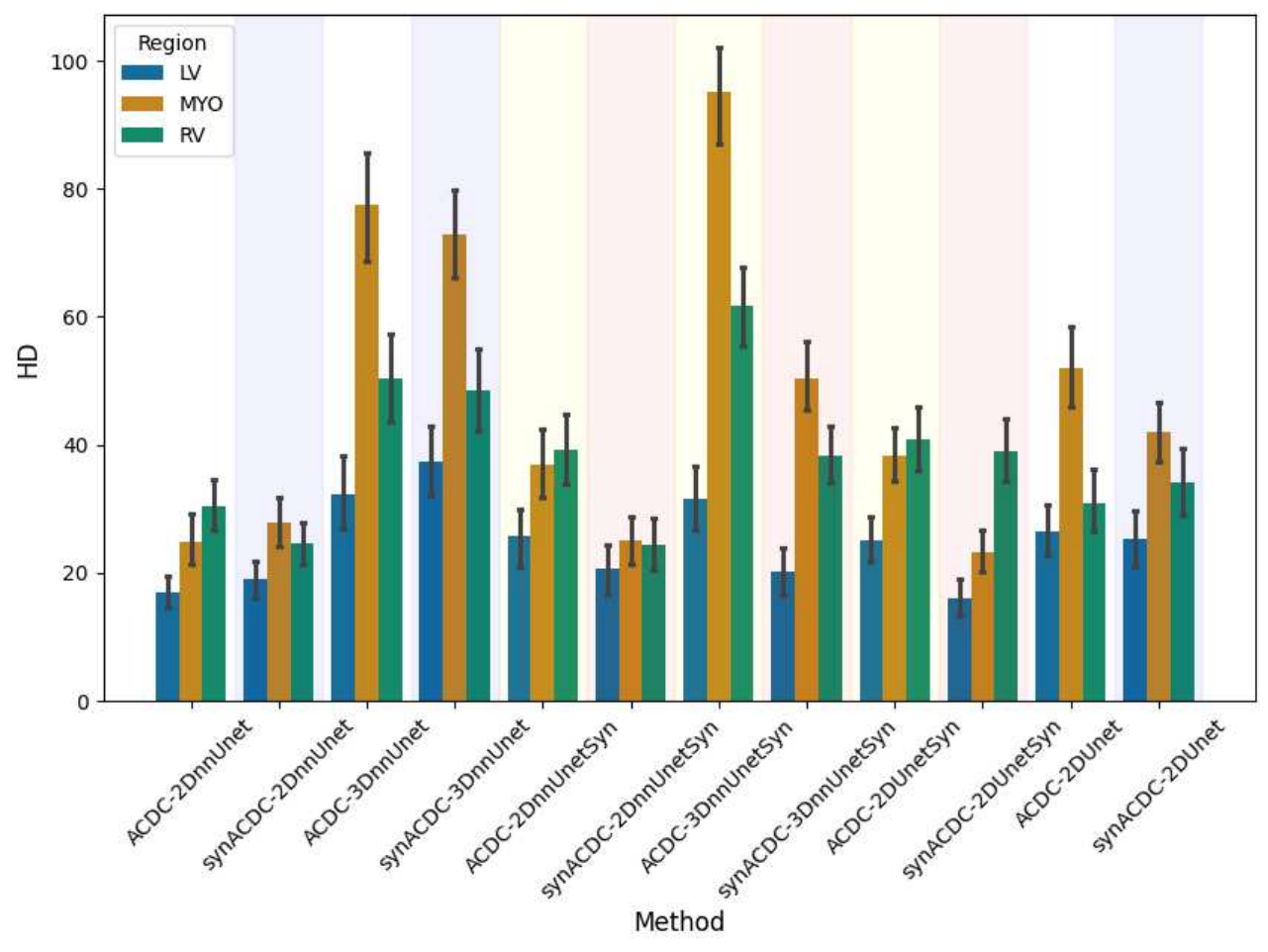}
        \caption{\textbf{Mean Hausdorff Distance}}
        \label{fig:hd_plot}
    \end{subfigure}
    \hfill
    \begin{subfigure}[b]{0.48\linewidth}
        \centering
        \includegraphics[width=\linewidth]{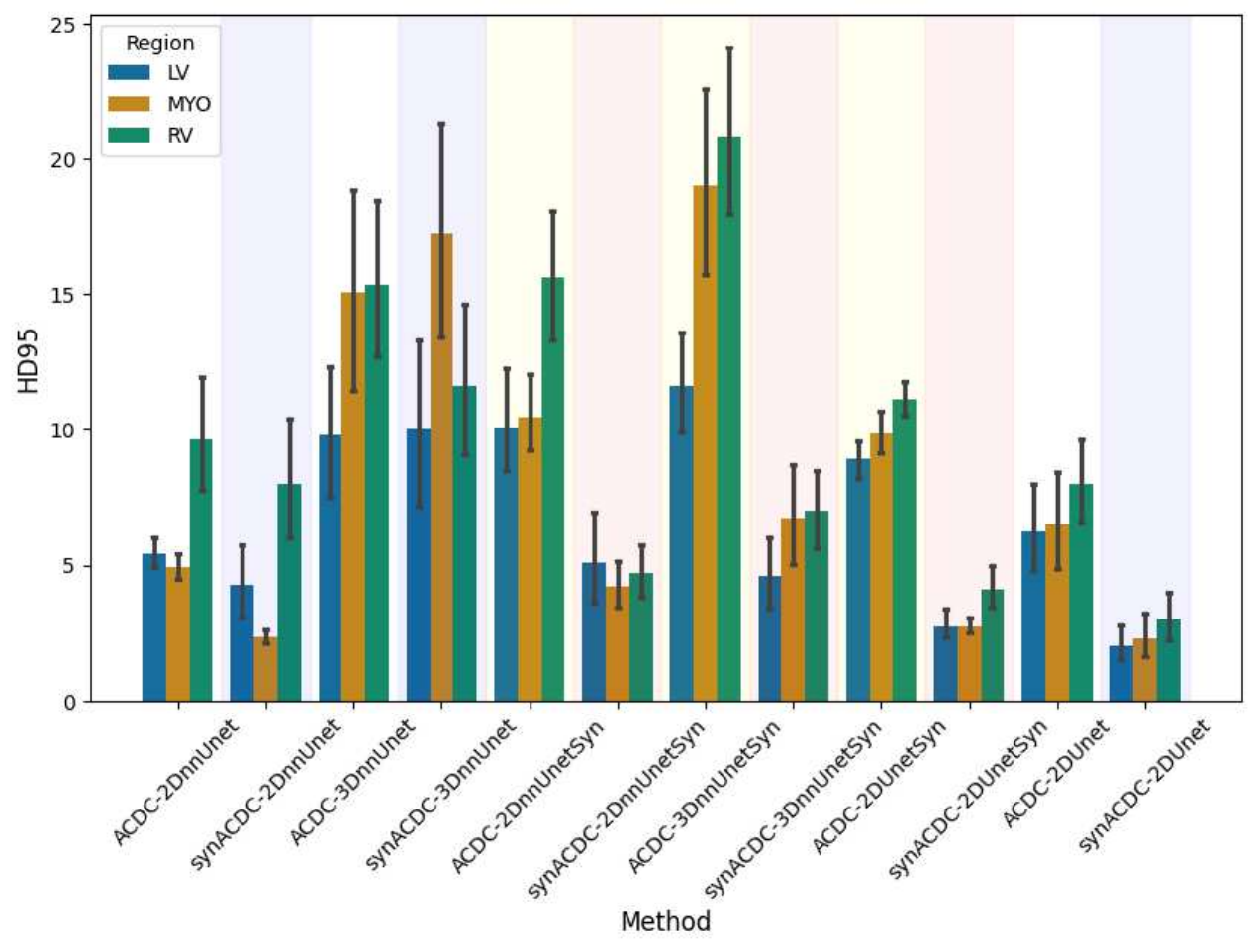}
        \caption{\textbf{Mean 95\% Hausdorff Distance}}
        \label{fig:hd95_plot}
    \end{subfigure}

    \caption{\textbf{Performance metrics with 95\% confidence intervals across anatomical structures.} Each bar chart summarizes segmentation performance per structure—Left Ventricle (LV), Myocardium (MYO), and Right Ventricle (RV)—measured using: (a) Dice Similarity Coefficient (DSC), (b) Intersection over Union (IoU), (c) Hausdorff Distance (HD), and (d) 95th percentile HD (HD95). Background shading indicates data types: faint blue for synthetic test data, faint yellow for synthetic source-trained models, and faint red for synthetic test data evaluated on synthetic-trained models.}
    \label{fig:performance_metrics}
\end{figure}

\end{document}